\documentclass[runningheads]{llncs}
\usepackage{graphicx}
\usepackage{amsmath,amssymb} % define this before the line numbering.
\usepackage{color}
\usepackage{float}
\usepackage{multirow}
\usepackage{tabularx}
\usepackage{soul}
\usepackage{algpseudocode}
\usepackage{threeparttable}
\usepackage{subfigure}
\usepackage{enumitem}
\usepackage{nicefrac}
\usepackage{cite}
\newcommand{\M}[1]{\mathcal{#1}}
\usepackage{bm}

\usepackage{array}
\newcolumntype{L}[1]{>{\raggedright\let\newline\\\arraybackslash}m{#1}}
\newcolumntype{C}[1]{>{\centering\let\newline  \\\arraybackslash}m{#1}}
\newcolumntype{R}[1]{>{\raggedleft\let\newline \\\arraybackslash}m{#1}}

\begin{document}
% \linenumbers
\pagestyle{headings}
\mainmatter
\def\ECCVSubNumber{4796}

\title{AutoSTR: Efficient Backbone Search for \\ Scene Text Recognition} % Replace with your title

\titlerunning{AutoSTR: Efficient Backbone Search for  Scene Text Recognition} 

\authorrunning{H. Zhang, Q. Yao, M. Yang, Y. Xu and X. Bai} 
\author{Hui Zhang$^{1,2}$, Quanming Yao$^{2}$, Mingkun Yang$^1$, Yongchao Xu$^1$, Xiang Bai$^{1}$}
\institute{$^1$Department of Electronics and Information Engineering, \\
	Huazhong University of Science and Technology \\
	$^2$4Paradigm Inc.}
%\author{Anonymous ECCV submission}
%\institute{Paper ID \ECCVSubNumber}

\maketitle

\begin{abstract}
Scene text recognition (STR) is challenging due to the diversity of text instances and the complexity of scenes.
However,
no STR methods can adapt backbones to different diversities and complexities.
%is an open issue that has not been explored.
%The community has paid increasing attention to boost the performance by improving the pre-processing image module, 
%like rectification and deblurring, or the sequence translator. 
%However, another critical module, i.e., 
%the feature sequence extractor, 
In this work, 
inspired by the success of neural architecture search (NAS),
%which can identify better architectures than human-designed ones,
we propose automated STR (AutoSTR),
which can address the above issue by searching data-dependent backbones.
Specifically,
we show both choices on operations and the downsampling path
are very important in the search space design of NAS.
Besides,
since 
no existing NAS algorithms can handle 
the spatial constraint on the path,
we propose a two-step search algorithm,
which decouples operations and downsampling path,
for an efficient search in the given space.
Experiments demonstrate that,
by searching data-dependent backbones,
AutoSTR can outperform the state-of-the-art approaches on standard benchmarks
with much fewer FLOPS and model parameters.
\footnote{Correspondence authors are Quanming Yao and Xiang Bai,
	and code is available at \url{https://github.com/AutoML-4Paradigm/AutoSTR.git}.}
\keywords{Scene Text Recognition, Neural Architecture Search, 
	Convolutional Neural Network, Automated Machine Learning}
\end{abstract}

\section{Introduction}\label{sec:intro}
\begin{figure}[t]
	\centering
	\includegraphics[width=0.99\textwidth]{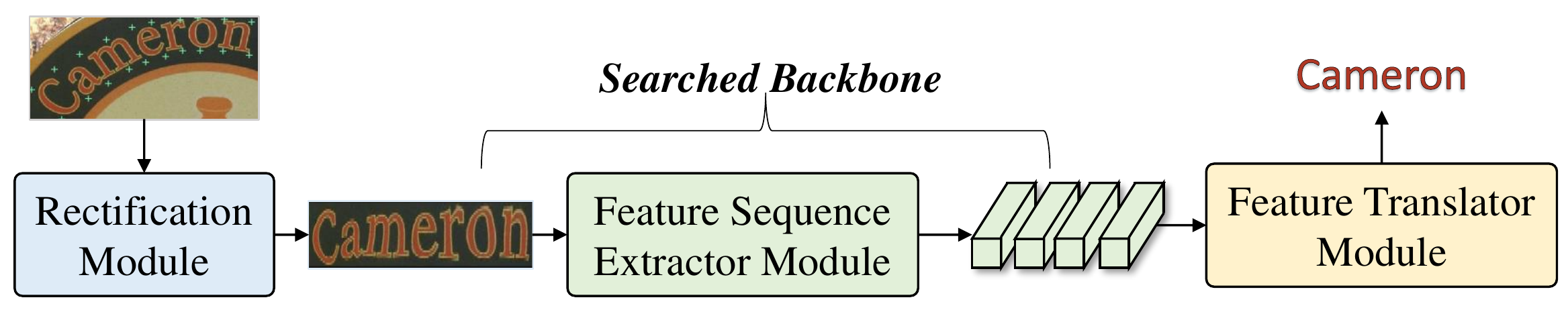}
	% \vspace{-20px}
	\caption{Illustration of general structure of text recognition pipeline. Feature sequence extractor is searched in this paper.}
	\label{fig:general_structure}
	% \vspace{-15px}
\end{figure}

Scene text recognition (STR) \cite{long2018scene,zhu2016scene}, 
which targets at recognizing text from natural images,
has attracted great interest from both industry and academia due to its huge commercial values in a wide range of applications, such as identity authentication, digital financial system, and vehicle license plate recognition \cite{xie2018robust,almazan2014word,dutta2018localizing}, 
etc. 
However, 
natural images are diverse,
the large variations in size, fonts, background and layout all
make STR still a very challenge problem \cite{shi2018aster}.
A STR pipeline (Fig.\ref{fig:general_structure}) \cite{shi2018aster,zhan2019esir} usually consists of three modules: 
a rectification module, 
which aims at rectifying irregular text image to a canonical form before recognition;
a feature sequence extractor, 
which employs a stack of convolutional layers to convert the input text image to a feature sequence;
and a feature translator module, which is adopted to translate the feature sequence into a character sequence.

In recent years, 
numerous methods~\cite{shi2018aster,yang2019symmetry,zhan2019esir} 
% \QM{you write ``decades'' and ``numerous'', but only with 3 papers in 2018-19}
have
successfully improved the text recognition accuracy via enhancing the performance of 
the rectification module. 
%This module has become a popular plug-in part to rectify irregular text image as regular one.
As for feature translator, inspired by some other sequence-to-sequence tasks, 
such as speech recognition~\cite{graves2006connectionist} and machine translation~\cite{bahdanau2014neural}, 
the translation module is also elaborately explored with both Connectionist Temporal Classification (CTC) based~\cite{shi2016end} and attention based methods~\cite{bai2018edit,cheng2017focusing,cheng2018aon,shi2018aster,yang2019symmetry}. 
In the contrast, the design of feature sequence extractor is relatively fewer explored.
% \QM{you can not say ``study'' \cite{baek2019wrong}.
	% you should say the designed of feat. seq. extr. is relatively fewer explored.}
How to design a better feature sequence extractor has not been well discussed in the STR literature.
However, 
text recognition performance 
% \footnote{+++ you can not say ``mainly''.
% 	it's annoying and not precise.
% 	1) yourself have listed many papers for stage A and C;
% 	2) VGG to ResNet can only show VGG is not a good choice;
% 	3) yourself do not have experiments supporting this claim}
can be greatly affected by feature sequence extractor. 
For example, the authors~\cite{shi2018aster} can obtain significant performance gains by simply replacing feature extractor from vgg~\cite{simonyan2014very} to ResNet~\cite{he2016deep}. 
Furthermore, the feature sequence extractor bears heavy calculation and storage burden~\cite{li2019show,yang2019symmetry}. 
Thus, 
no matter for effectiveness or efficiency, the architecture of feature sequence extractor should be paid more attention to.

Besides, 
neural architecture search (NAS) \cite{elsken2019neural,yao2018taking} has also made a 
great success in designing data-dependent network architectures,
of which the performances exceed the architectures crafted by human experts in many computer vision tasks,
e.g., 
image classification~\cite{DARTS,pham2018enas},
semantic segmentation~\cite{liu2019auto} and object detection~\cite{chen2019detnas}.
Thus, 
rather than adopt an off-the-shelf feature extractor (like ResNet~\cite{he2016deep}) from other tasks,
a data-dependent architecture should be redesigned for a better text recognition performance.

In this paper, we present the first work,
i.e.,
AutoSTR, 
on searching feature sequence extractor (backbone) 
for STR. 
First we design a domain-specific search space for STR, 
which contains both choices on operation for every convolution layer and constraints feature downsampling path.
% \QM{1. novelty in space design; then 2. novelty in algorithm design;
% 	see abs.}
Since no existing NAS algorithms
can handle the path constraint efficiently,
%And different from the existing NAS algorithms~\cite{Cai19Proxylessnas,chen2019detnas,liu2019auto,DARTS}, 
we propose
% \QM{AutoSTR is never used previously, strange to appear without a justification} 
a novel two-step search pipeline, which decouples operation and downsampling path search.
% first searches a downsampling strategy and then searches operation for each convolutional layer.
% But most of existing NAS methods~\cite{} only focus on finding appropriate types of convolutional layers. 
% Only AutoDeepLab\cite{Liu2019AutoDeepLab} searches downsampling strategy in the segmentation task. 
% But it is not suitable for our search space due to the use of special 
% \footnote{+++ you can guess: how many people can understand what your ``Viterbi'' mean here.}
% Viterbi decoding. 
% Therefore, we propose a an efficient two-way search algorithm, called AutoSTR, to simultaneously search the types of convolution and downsampling path in our search space. 
By optimizing the recognition loss with complexity regularization, we achieve a trade off between model complexity and recognition accuracy. Elaborate experiments demonstrate that, given a general text recognition pipeline, the searched sequence feature extractor can achieve state-of-the-art results with fewer FLOPS and parameters.  
The main contributions of this paper are summarized as follows:
\begin{itemize}[leftmargin=*]
\item We discover that the architecture of the feature extractor,
which is of great importance to STR,
has not been well explored in the literature.
This motivates us to design a data-dependent backbone to boost text recognition performance,
which is also the first attempt to introduce NAS into STR.

\item 
We introduce a domain-specific search space for STR, 
which contains choices for downsampling path and operations.
Then
we propose a new search algorithm,
which decouples operations and downsampling path for an efficient search in the space. 
We further 
incorporate an extra regularizer into the searching process,
which helps effectively trade off 
the recognition accuracy with the model size.

\item 
Finally,
extensive experiments are carried on various benchmark datasets.
Results demonstrate that,
AutoSTR can discover data-dependent backbones,
and outperform the state-of-the-art approaches 
with much fewer FLOPS and model parameters.

\end{itemize}

\section{Related Works}

\subsection{Scene Text Recognition (STR)}
As in Sec~\ref{sec:intro},
the general pipline (Fig.\ref{fig:general_structure}) of sequence-based STR methods \cite{shi2018aster,zhan2019esir},
where a rectification module,
a feature sequence extractor module
and 
a feature translator module are involved.
Currently, most works focus on improving rectification module or feature translator. 
Shi \textit{et al.} \cite{shi2016robust,shi2018aster} first introduce spatial transform network (STN)\cite{jaderberg2015spatial} for rectifying irregular text to a canonical form before recognition. 
Since then, \cite{luo2019multi,yang2019symmetry,zhan2019esir} further push it forward and make the rectification module become a plug-in part. 
As for feature translator, CTC based and attention based decoder dominate this landscape for a long time~\cite{shi2016end,shi2018aster,bai2018edit}. 
Nevertheless, as another indispensable part, the feature sequence extractor has not been well discussed in the recent STR literature. 
As shown in 
Tab.\ref{tab:txtreg}, 
although different feature extractors are used in \cite{shi2016end,yang2019symmetry}, 
they just follow the architecture proposed for other fundamental tasks, like image classification. 
But recently, some authors observe that the architectures of feature extractor have gaps between different tasks, 
like semantic segmentation~\cite{liu2019auto} and object detection~\cite{chen2019detnas}. 
Therefore, 
these popular feature extractor might not be perfectly suitable for STR,
and it is important to search a data-dependent architecture.

\begin{table}[t]
	\centering
	\caption{Comparison with example text recognition and NAS methods
		that contain a downsampling path search. 
		``feat.seq.extractor'',
		``DS", ``seq. rec.", ``cls.",``seg." and ``grad. desc." means feature sequence extractor,
		downsampling, sequence recognition, classification,segmentation and gradient descent algorithm respectively.}
	\setlength\tabcolsep{3pt}
	\begin{tabular}{c|c | C{55px} C{50px} | C{45px} | C{50px}}
		\hline
		&        \multirow{2}{*}{\textbf{model}}         & \multicolumn{2}{c|}{\textbf{feat. seq. extractor}} & \multirow{2}{*}{\textbf{task}} & \textbf{search}      \\
		&                                       & \textbf{operation}   & \textbf{DS path }                       &                       & algorithm   \\ \hline
		hand-    &        CRNN~\cite{shi2016end}         & vgg      & fixed                          & seq. rec.             & ---         \\ \cline{2-6}
		designed &       ASTER~\cite{shi2018aster}       & residual & fixed                          & seq. rec.             & ---         \\ \cline{2-6}
		         &     SCRN~\cite{yang2019symmetry}      & residual & fixed                          & seq. rec.             & ---         \\ \cline{2-6}\hline
		NAS    	 &     DARTS~\cite{DARTS}				 & searched & fixed							 & cls.					 & grad. desc. \\ \cline{2-6}
			 	 &    AutoDeepLab~\cite{liu2019auto}     & searched & one-dim                        & seg.                  & grad. desc. \\ \cline{2-6}
		         &                AutoSTR                & searched & two-dim                        & seq. rec.             & two-step    \\ \hline
	\end{tabular}
	\label{tab:txtreg}
\end{table}

\subsection{Neural Architecture Search (NAS)} \label{sec:nas}
Generally,
there are two important aspects in neural architecture search (NAS) \cite{elsken2019neural,yao2018taking}, i.e.,
the \textit{search space} and \textit{search algorithm}.
The search space defines
possibilities of all candidate architectures,
and the search algorithm attempts to find suitable architectures from the designed space.
Specifically,
a proper space should explore domain information and cover good candidates,
which are designed by humans.
Besides,
it cannot be too large, 
otherwise, 
no algorithms can efficiently find good architectures.
Thus,
usually designing a better search space by domain knowledge can make the searching process easier.

Classically, network architectures are treated as hyper-parameters,
which are optimized by an algorithm like reinforcement learning~\cite{zoph2016neural} and evolution algorithms~\cite{xie2017genetic}. 
These methods are expensive since they need to train each sampled architecture fully. 
Currently one-shot neural architecture search (OAS)
\cite{Cai19Proxylessnas,DARTS,pham2018enas,yao2019differentiable} 
have significantly reduced search time by sharing the network weights during the search progress. 
Specifically,
these methods encode the whole search space into a supernet, which consists of all candidate architectures.
Then, 
instead of training independent weights for each architecture, 
distinctive architecture inherits weights from the supernet. 
In this way, architectures can be searched by training the supernet once, 
which makes NAS much faster. 
However,
as the supernet is a representation of the search space,
a proper design of it is a non-trivial task.
Without careful exploitation of domain information,
OAS methods may not be even better than random search~\cite{li2019random,sciuto2019evaluating}.

\section{Methodology}

%As an indispensable and general module, feature sequence extractor play a crucial role for most current text recognition methods \cite{shi2018aster,shi2016end,bai2018edit,yang2019symmetry}. 
%Our goal aims at extending NAS to search a better backbone for the general text recognition pipeline. 
%In this section, we first formally introduce the problem definition to automatically search network architecture for STR in Sec~\ref{problem_formulation}. Then, we describe our novel search space and search algorithm in Sec~\ref{search_space} and \ref{sec:search_alg} respectively.

\subsection{Problem Formulation}\label{problem_formulation}
As the input of STR is natural images,
feature sequence extractor is constructed with convolutional layers. 
%owing to its power to extract discriminative features \cite{he2016deep}. 
A convolutional layer $\M{C}$ can be defined as $\M{C}(X; o, s^h, s^w)$ (hereinafter referred to as $\M{C}(X)$),
where $X$ is input tensor, 
%$\M{C}$ is a convolutional layer with hyper-parameters $o$, $s^h$ and $s^w$. 
%Specifically, 
$o$ denotes convolution type, 
(e.g., 3$\times$3 convolution, 
5$\times$5 depth-wise separable convolution),
$s^h$ and $s^w$ represent its stride in height and width direction respectively. 
Therefore, a backbone $\M{N}$ can be regarded as a stack of $L$ convolution layers, 
i.e., $\M{C}_L(\dots\M{C}_2(\M{C}_1(X)))$. 
After been processed by $\M{N}$, 
$X$ with input size $(H, W)$ will be mapped into a feature map with 
a fixed size output to the feature translator module.

To determine a data-dependent backbone for STR,
we need to search proper architectures, 
which is controlled by $\mathcal{S} \equiv \{(s^h_i, s^w_i)\}_{i = 1}^L$ (for strides)
and $\mathcal{O} \equiv \{ o_i \}_{i = 1}^L$ (for convolution type). 
Let $\mathcal{L}_{\text{tra}}$ measure the loss of the network on training dataset
and $\mathcal{A}_{\text{val}}$ measure the quality of architecture on the validation set.
% and $\mathcal{A}_{\text{val}}$ be a differentiable criterion
% measuring the quality of the architecture.
%(resp. validation performance). 
%$\M{A}_{val}$ means the performance of $\M{N}$ on the validation dataset.
We formulate 
the AutoSTR  problem as:
\begin{align}
\min_{\M{S}, \M{O}}
\mathcal{A}_{\text{val}}(\M{N}(\bm{w}^*, \M{S}, \M{O})),
\quad
\text{s.t.}
\quad
\begin{cases}
\bm{w}^* 
= \arg\min_{\bm{w}}
\mathcal{L}_{\text{tra}}(\M{N}(\bm{w}, \M{S}, \M{O})) 
\\ 
\M{S} \in \M{P}
\end{cases},
\label{eq:opt}
\end{align}  
where $\M{S}$ and $\M{O}$ are upper-level variables representing architectures, 
and $\bm{w}$ as lower-level variable and $\M{P}$ as constraint,
i.e.,
\begin{equation*}
\mathcal{P}
\equiv
\{
\{ (s^h_i, s^w_i) \}_{i = 1}^L
\;|\;
\nicefrac{H}{\prod_{i=1}^{L}s^h_i} = c_1, 
\nicefrac{W}{\prod_{i=1}^{L}s^w_i} = c_2
\}.
\end{equation*}
Specifically,
$c_1$ and $c_2$ are two application dependent constants;
and the constraint 
$\mathcal{P}$ is to ensure
the output size of 
the searched backbone is aligned with
the input size of the subsequent feature translator module.

\subsection{Search Space}
\label{search_space}

As explained in Sec~\ref{sec:nas},
the search space design is a key for NAS.
Here,
we design a two-level hierarchical search space for the AutoSTR problem in \eqref{eq:opt},
i.e., the downsampling-path level and operation level, to represent the selection range of $\M{S}$ and $\M{O}$ respectively.

\subsubsection{Downsampling-path level search space.}\label{sec:path_search_space}

Since the characters are horizontally placed in the rectified text image, 
following \cite{shi2016end},
we use CNN to exact a feature sequence to represent them. 
To reserve more discriminable features of the characters in compact text or in narrow shapes, 
a common way is to keep collapsing along the height axis until it reduces to 1, 
but compress less along the width axis to ensure that the length of final sequence is greater 
than the length of characters~\cite{shi2016end,shi2018aster,zhan2019esir}. 
Specifically, the current mainstream methods are using the feature extractor proposed in 
ASTER~\cite{shi2018aster}. The height of the input text image is unified to a fixed size, like 32. 
And to reserve more resolution along the horizontal axis in order to distinguish neighbor characters~\cite{bai2018edit,shi2016end,shi2018aster,yang2019symmetry,zhan2019esir}, 
the strides hyper-parameter $s$ only selected from $\{(2, 2), (2, 1), (1, 1)\}$, where $(2, 2)$ appears twice and $(2, 1)$ appears three times in the whole downsampling path to satisfy $\M{P}$ with $\prod_{i=1}^{L}s^h_i = 32$ and $\prod_{i=1}^{L}s^w_i = 4$. Finally, a text image with size $(32, W)$ is mapped into a feature sequence with a length of $\nicefrac{W}{4}$. 

\begin{figure}[t]
	\centering
	\includegraphics[width = 0.99\textwidth]{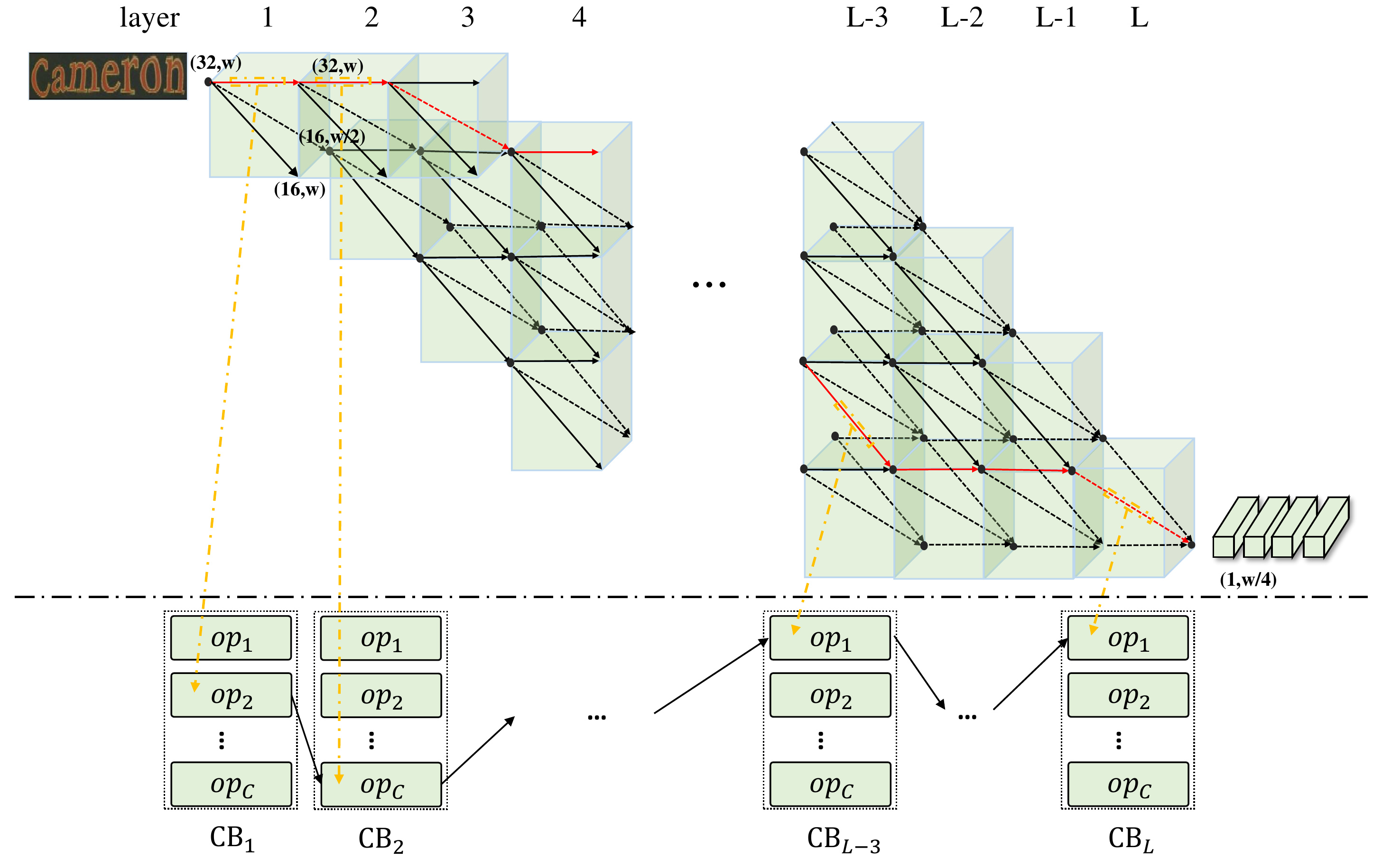}
	\caption{Search space illustration.
		Top: a 3D-mesh representation of the downsampling-path level search space.
		Bottom: operation level search space, where 
		``CB$_i$'' donates the $i$th \textit{choice block}
		and each block allows $C$ choices for operations.}
	\label{fig:path_level}
\end{figure}

The downsampling-path level search space is illustrated in Fig.\ref{fig:path_level}.
Our goal is to find an optimal path in this 3D-mesh, and the downsampling strategy in \cite{shi2018aster} is a spatial case in this search space. 
To the best of our knowledge, 
there are no suitable methods to search in such a constrained search space.

\subsubsection{Operation level search space.} 
The convolutional layers of current text recognition networks usually share the same operation \cite{cheng2017focusing,shi2018aster,zhan2019esir}, 
such as $3 \times 3$ residual convolution. 
% \footnote{+++ what if we use operations in ASTER?
% 	how is the performance?
% 	I mean the searched path + residual convlolution}
Instead of setting each $o_i$ to be a fixed operation, 
we select a operator for each convolutional layer from a \textit{choice block} with $C$ parallel operators, 
as illustrated in the bottom of Fig.\ref{fig:path_level}. 
Then, we can obtain a deep convolutional network by stacking these \textit{choice blocks}. 
Our choices on operations are inspired by MobileNetV2~\cite{sandler2018mobilenetv2}, 
which uses lightweight depthwise convolutions to save FLOPS and the number of parameters. 
Thus, 
we build a set of mobile inverted bottleneck convolution layers (MBConv) with various kernel sizes $k \in \{3,5\}$, 
expansion factors  $e \in \{1,3,6\}$ and a skip-connect layer. 

%\begin{figure}[t]
%	\centering
%	
%	\caption{}
%	
%\end{figure}

\begin{table}[ht]
	\centering
	\caption{Basic operation (i.e., op$_i$'s) in the choice block (the bottom of Fig.\ref{fig:path_level}), 
		where ``k'' denotes kernel size and ``e'' denotes expansion factors. }
	\setlength\tabcolsep{2pt}
	\begin{tabular}{c | c | c | c}
		\hline
		MBConv(k:3$\times$3,e:1) & MBConv(k:3$\times$3,e:3) & MBConv(k:3$\times$3,e:6) & MBConv(k:5$\times$5,e:1) \\ \hline
		MBConv(k:5$\times$5,e:3) & MBConv(k:5$\times$5,e:6) &    \multicolumn{2}{c}{Skip-Connect}              \\ \hline
	\end{tabular}
	\label{tab:choice_block}
	% \vspace{-10px}
\end{table}

\subsubsection{Complexity of the search space.}
When $L=15$,
there are 30030 possible downsampling paths in search space illustrated in Fig.\ref{fig:path_level}.
On operation level,
if we allow it to be one of the seven operations in Tab.\ref{tab:choice_block},
then it leads to a total number $30030 \times 7^{15} \simeq 1.43 \times {10}^{17}$ possible architectures for the backbone,
which is prohibitively large.
In the sequel,
we show a two-step 
algorithm which can efficiently search through the space.

\subsection{Search Algorithm}\label{sec:search_alg}

Since the combination of 
$\M{S}$ and $\M{O}$ generates a very space,
directly optimizing the problem in \eqref{eq:opt} is a huge challenge. 
Motivated by the empirical observation
that choices of the downsampling path and the operations
are almost orthogonal with each other (see Sec~\ref{sec:abl}),
in this section,
we decouple the process of searching $\M{S}$ and $\M{O}$ into two steps for the backbone search.
Specifically,
in the first step, 
we fix the operation $\M{O}$ as the 3$\times$3 residual convolution (denote as $\hat{\mathcal{O}}$)
and search for the downsampling path. 
In the second step, we search a convolution operation for every layer in the path.

\subsubsection{Step 1: search downsampling path.}
Let $\M{L}_{\text{rec}}(\M{N}; \M{D})$ measure the sequence cross-entropy loss \cite{shi2018aster}
with predictions from a network $\M{N}$ on a dataset $\M{D}$,
and $\M{D}_{\text{tra}}$ (resp. $\M{D}_{\text{val}}$) denotes the training (resp. validation) dataset.
In this step,
we search downsampling path when operations are fixed,
and 
\eqref{eq:opt} becomes
\begin{align}
\M{S}^*
& =
\arg\min_{\M{S} \in \M{P}}
\M{L}_{\text{rec}}(\M{N}(\bm{w}^*, \M{S}, \hat{\M{O}}); \M{D}_{\text{val}}),
\label{eq:stage1}
\\
\text{\;s.t.\;}
\bm{w}^* 
& = \arg\min_{\bm{w}}
\mathcal{L}_{\text{rec}}(\M{N}(\bm{w}, \M{S}, \hat{\M{O}}); \M{D}_{\text{tra}}).
\notag
\end{align}
%We distribute the downsampling position uniformly in the network, i.e., 
%each convolutional stage has the same number of convolutional layers, which is straight and effective. 
As in Sec~\ref{sec:path_search_space}, 
downsampling path can only have two $(2, 2)$ and three $(2, 1)$ to satisfy $\M{P}$, 
which are denoted as downsampling type $A$ and $B$ respectively. 
Note that some current NAS methods~\cite{DARTS,yao2019differentiable} use the same number of layers per convolution stage and have achieved good results. Using this reasonable prior knowledge, for a network with $L=15$, we set downsampling at layers 1, 4, 7, 10, 13, and equally separate the network into five stages. 
Then the downsampling strategies can be divided into 10 types of typical paths: 
\textsf{AABBB}, 
\textsf{ABABB}, 
\textsf{ABBAB}, 
\textsf{ABBBA}, 
\textsf{BAABB}, 
\textsf{BABAB}, 
\textsf{BABBA}, 
\textsf{BBAAB}, 
\textsf{BBABA}, and 
\textsf{BBBAA}. 
We can do a small grid search in these typical paths to find a good path that is close to $\M{S}^*$. Then by learning the skip-connect in searching step 2, we can reduce the number of layers for each convolutional stage.

\subsubsection{Step 2: search operations.}\label{sec:search_op}
First,
inspired by recent NAS methods~\cite{Cai19Proxylessnas,guo2019single} 
we associate the operation $\text{op}_i^l$ at the $l$th layer with a hyperparameter $\alpha_i^l$,
which relaxes the categorical choice of a particular operation 
in the \textit{choice block} (Fig.\ref{fig:path_level}) to be continuous. 
Since 
$\text{op}_i^l$'s influence 
both complexity and performance of the backbone~\cite{sandler2018mobilenetv2,Cai19Proxylessnas,yao2019differentiable},
we introduce a regularizer,
i.e., 
%\footnote{+qm+ the range of log is not clear.}
\begin{align} \label{eq:reg}
r(\bm{\alpha}) = 
[  \log ( \sum_{l = 1}^L \sum_{j = 1}^{C} \text{FLOPS}(\text{op}^l_j) )  \cdot \alpha_{j}^l / \log \M{G} ]^{\beta},
\end{align}
on the architecture $\bm{\alpha}$,
where $\beta > 0$ and $\M{G} > 1$ be application-specific constants.
Then,
\eqref{eq:opt} is transformed as
\begin{align}
\bm{\alpha}^* 
& = 
\arg
\min_{\bm{\alpha}} 
r(\bm{\alpha}) \cdot \M{L}_{\text{rec}} ( \M{N}(\bm{w}^*, \M{S}^*, \bm{\alpha}), \M{D}_{\text{val}} ),
\label{eq:stage2}
\\
\text{\;s.t.\;}
\bm{w}^* 
& = \arg\min_{\bm{w}}
\mathcal{L}_{\text{rec}}(\M{N}(\bm{w}, \M{S}^*, \bm{\alpha}); \M{D}_{\text{tra}}).
\notag
\end{align}
We can see that 
when $\M{L}_{\text{rec}}$ is the same,
$r(\bm{\alpha})$ make architectures with less FLOPS favored.
Thus,
the regularizer can effectively trade off the accuracy with the model size.
Finally,
\eqref{eq:stage2} can be solved 
by many existing NAS algorithms,
such as 
DARTS~\cite{DARTS},
ProxylessNAS~\cite{Cai19Proxylessnas} and NASP~\cite{yao2019differentiable}.
Here,
we adopt
ProxylessNAS~\cite{Cai19Proxylessnas} as it consumes
the GPU memory less.

\subsection{Comparison with other NAS works}
\label{exist_nas}

The search constraint $\mathcal{P}$ on the downsampling path for is new to NAS,
which is specific to STR.
Unfortunately,
none of existing NAS algorithms can effectively deal with the
AutoSTR problem in \eqref{eq:opt} here.
It is the proposed search space (Sec~\ref{search_space}) and two-stage search algorithm
(Sec~\ref{sec:search_alg}) that
make NAS for STR possible (see Tab.\ref{tab:txtreg}).
We notice that AutoDeepLab~\cite{liu2019auto} also considers the downsampling path.
However, it targets at in segmentation task and
its unique decoding method is not suitable for our search space. 

\section{Experiments}

\subsection{Datasets}
We evaluate our searched architecture on the following benchmarks that are designed for general STR.
Note that
the images in the first four datasets are regular
while others are irregular. 
\begin{itemize}[leftmargin=*]
\item IIIT 5K-Words (\textbf{IIIT5K})~\cite{mishra2012top}:
it contains 5,000 cropped word images for STR, 2,000 for validation and other 3,000 images for testing;
all images are collected from the web.

\item Street View Text (\textbf{SVT})~\cite{wang2011end}:
It is harvested from Google Street View. 
Its test set contains 647 word images,
which exhibits high variability and often has low resolution. 
Its validation set contains 257 word images.

\item ICDAR 2003 (\textbf{IC03})~\cite{lucas2005icdar}:
it contains 251 full scene text images. Following~\cite{wang2011end}, we discard the test images containing non-alphanumeric characters or have less than three characters. The resulting dataset contains 867 cropped images for testing and 1,327 images as validation dataset.

\item ICDAR 2013 (\textbf{IC13})~\cite{karatzas2013icdar}:
it contains 1,015 cropped text images as test set and 844 images as validation set.

\item ICDAR 2015 (\textbf{IC15}):
it is the 4th Challenge in the ICDAR 2015 Robust Reading Competition~\cite{karatzas2015icdar}, 
which is collected via Google Glasses without careful positioning and focusing. 
As a result, 
the dataset consists of a large proportion of blurred and multi-oriented images. 
It contains 1811 cropped images for testing and 4468 cropped text images as validation set.

\item SVT-Perspective (\textbf{SVTP}):
it is proposed in~\cite{quy2013recognizing} which targets for evaluating the performance of recognizing perspective text and contains 645 test images. 
Samples in the dataset are selected from side-view images in Google Street View. 
Consequently, a large proportion of images in the datasets are heavily deformed by perspective distortion.
\end{itemize}

\subsection{Implementation Details}

The proposed method is implemented in PyTorch. We adopt ADADELTA~\cite{zeiler2012adadelta} with default hyper-parameters (rho=0.9, eps=1e-6, weight decay=0) to minimize the objective function. 
When searching the downsampling path, 
we train 5 epochs for each typical path where the convolutional layers are equipped with default $3\times3$ convolution. 
In the operation searching step, we warm-up weights of each choice block by uniformly selecting operations for one epoch. 
And then use the method proposed in Sec~\ref{sec:search_op} to jointly train architecture parameters and weight parameters for two epochs. 
In the evaluation step, the searched architectures are trained on Synth90K~\cite{jaderberg2014synthetic} and SynthText~\cite{gupta2016synthetic} from scratch without finetuning on any real datasets. 
All models are trained on 8 NVIDIA 2080 graphics cards.
Details of each module in Fig.\ref{fig:general_structure} are as follows:

\subsubsection{Rectification module.} 
Following~\cite{shi2018aster}, we use Spatial Transformer Network (STN) to rectify the input text image. The STN contains three parts: (1) Localization Network, which consists of six 3$\times$3 convolutional layers and two fully connected layers. In this phase, 20 control points are predicted. Before fed into the localization network, the input image is resized to 32$\times$64. (2) Grid Generator, which yields a sampling grid according to the control points. (3) Sampler, which generates a rectified text image with the sampling grid. The sampler produces a rectified image of size 32$\times$100 from the input image of size 64$\times$256 and sends it to the subsequent sequence feature extractor.

\subsubsection{Feature sequence extractor module.} 
The feature sequence extractor dynamically changes during the search phase. For every test dataset, an architecture is searched. During the search phase, Synth90K~\cite{jaderberg2014synthetic} (90k) and SynthText~\cite{gupta2016synthetic} (ST) are used for training. The validation set of each target dataset is considered as the search validation set and is used to optimize the network structure. 
In order to prevent overfitting caused by too small validation set, 
we add extra COCO-Text training set to the search validation set.
% \QM{any refs?}. 
To control the complexity of the backbone, 
we only search those with less than 450M FLOPS which is close to 
that of the backbone used in ASTER. 
%Therefore, only AABBB and ABABB meet the requirement.
% \QM{any supporting refs for this number}. 
The maximum depth of our search network is 16 blocks, including a stem with 3$\times$3 residual convolution and 15 choice blocks. 
It has a total of 5 convolutional stages, each stage has 3 choice blocks, 
the number filters of each stage are respectively 32, 64, 128, 256, 512 respectively.

\subsubsection{Feature translator module.} 
A common attention based decoder~\cite{shi2018aster} is employed to translate the feature sequence to a character sequence. 
For simplicity, only left-to-right decoder is used. 
The module contains two layers of Bidirectional LSTM (BiLSTM) encoder (512 input units, 
256 hidden units) and an attention based GRU Cell decoder (1 layer, 
512 input units, 512 hidden units, 
512 attention units) to model variable-length character sequence. 
The decoder yields 95 character categories at each step, including digits, 
upper-case and lower-case letters, 32 ASCII punctuation marks and an end-of-sequence symbol (EOS).

\subsection{Comparison with State of the Art}

\begin{table}[t]
	\caption{Performance comparison on regular and irregular scene text datasets. ``ST", ``90k" are the training data of SynthText\cite{liu2018synthetically}, Synth90k\cite{jaderberg2014synthetic}, and ``extra" means extra real or synthetic data, respectively. The methods marked with ``$\dagger$" use the character box annotations. 
		``ASTER (ours)" is the reproduced ASTER baseline method, whose difference is that only left-to-right translator is equipped.}
	\label{tab:main_results}
	\centering
	\setlength\tabcolsep{4pt}
	\begin{threeparttable}
		\begin{tabular}{c|c|cccc|cc}
			\hline
			\multirow{2}{*}{\textbf{Methods}}    & \multirow{2}{*}{\textbf{Data}} &             \multicolumn{4}{c|}{\textbf{Regular}}              &    \multicolumn{2}{c}{\textbf{Irregular}}    \\ \cline{3-8}
			&                                & \textbf{IIIT5K} & \textbf{SVT} & \textbf{IC03} & \textbf{IC13} & \textbf{SVTP} & \textbf{IC15} \\ \hline
			Jaderberg et al.~\cite{jaderberg2016reading}		& 90k         & -    & 80.7 & 93.1 & 90.8 & -      & -     \\
			CRNN~\cite{shi2016end}							& 90k         & 81.2 & 82.7 & 91.9 & 89.6 & -      & -     \\
			RARE~\cite{shi2016robust}						& 90k  		  & 81.9 & 81.9 & 90.1 & 88.6 & 71.8 & -     \\
			$R^2$AM\cite{lee2016recursive}					& 90k  		  & 78.4 & 80.7 & 88.7 & 90.0 & -    & -       \\
			Yang et al.\cite{yang2017learning}              & 90k         & -    & -    & -    & -    & 75.8 & -     \\
			Char-net\cite{liu2018char}						& 90k         & 83.6 & 84.4 & 91.5 & 90.8 & 73.5    & -     \\
			Liu et al~\cite{liu2018synthetically}  			& 90k         & 89.4 & 87.1 & \underline{94.7} & 94.0 & 73.9 & - 	\\
			AON~\cite{cheng2018aon}      					& ST+90k      & 87.0 & 82.8 & 91.5 & -    & 73.0 & 68.2  \\
			FAN$^\dagger$~\cite{cheng2017focusing} 		    & ST+90k 	  & 87.4 & 85.9 & 94.2 & 93.3 & -      & 70.6  \\
			EP~\cite{bai2018edit}        					& ST+90k 	  & 88.3 & 87.5 & 94.6 & \textbf{94.4} & -      & 73.9  \\
			SAR~\cite{li2019show}    						& ST+90k& 91.5 & 84.5 & -    & 91.0 & 76.4 & 69.2 	\\
			CA-FCN$^\dagger$~\cite{liao2019scene} 		    & ST+extra    & 92.0 & 86.4 & -    & 91.5 & -    & - 	\\
			ESIR~\cite{zhan2019esir} 						& ST+90k      & 93.3 & \underline{90.2} & -    & 91.3 & 79.6  & 76.9  \\
			SCRN$^\dagger$~\cite{yang2019symmetry} 			& ST+90k 	  & \underline{94.4} & 88.9 & \textbf{95.0} & 93.9 &\underline{80.8}&\underline{78.7}  \\ 
			ASTER~\cite{shi2018aster}  					    & ST+90k      & 93.4 & 89.5 & 94.5 & 91.8 & 78.5 & 76.1 	\\
			\hline\hline
			ASTER (ours) 					    				& ST+90k      & 93.3 & 89.0   & 92.4 & 91.5 & 79.7 & 78.5 	\\
			AutoSTR 										& ST+90k      & \textbf{94.7} & \textbf{90.9} & 93.3 & \underline{94.2} & \textbf{81.7}  & \textbf{81.8} 	\\ \hline
		\end{tabular}
	\end{threeparttable}
	% \vspace{-10px}
\end{table}

\subsubsection{Recognition accuracy.}
Following~\cite{baek2019wrong}, all related works are compared in the unconstrained-lexicon setting. 
Equipped with the searched backbone, the whole framework is compared with other state-of-the-art methods, as shown in Tab.\ref{tab:main_results}. 
AutoSTR achieves the best performance in IIIT5K, SVT, IC15, SVTP and get comparable results in IC03, IC13. 
It is worth noting that AutoSTR outperforms ASTER (ours) on IIIT5K, SVT, IC03, IC13, SVTP, 
IC15 by 1.4\%, 1.9\%, 0.9\%, 2.7\%, 2\%, 3.3\%, which domonstrate the effectiveness of AutoSTR. Although SCRN can achieve comparable performance with AutoSTR, its rectification module requires extra character-level annotations for more precise rectification. As a plug-in part, AutoSTR is expected to further improve the performance while been equipped with the rectification module of SCRN.

\subsubsection{Memory and FLOPS.}
\begin{figure}[ht]
\centering
\subfigure[FLOPS (million).]{\includegraphics[width = 0.460\textwidth]{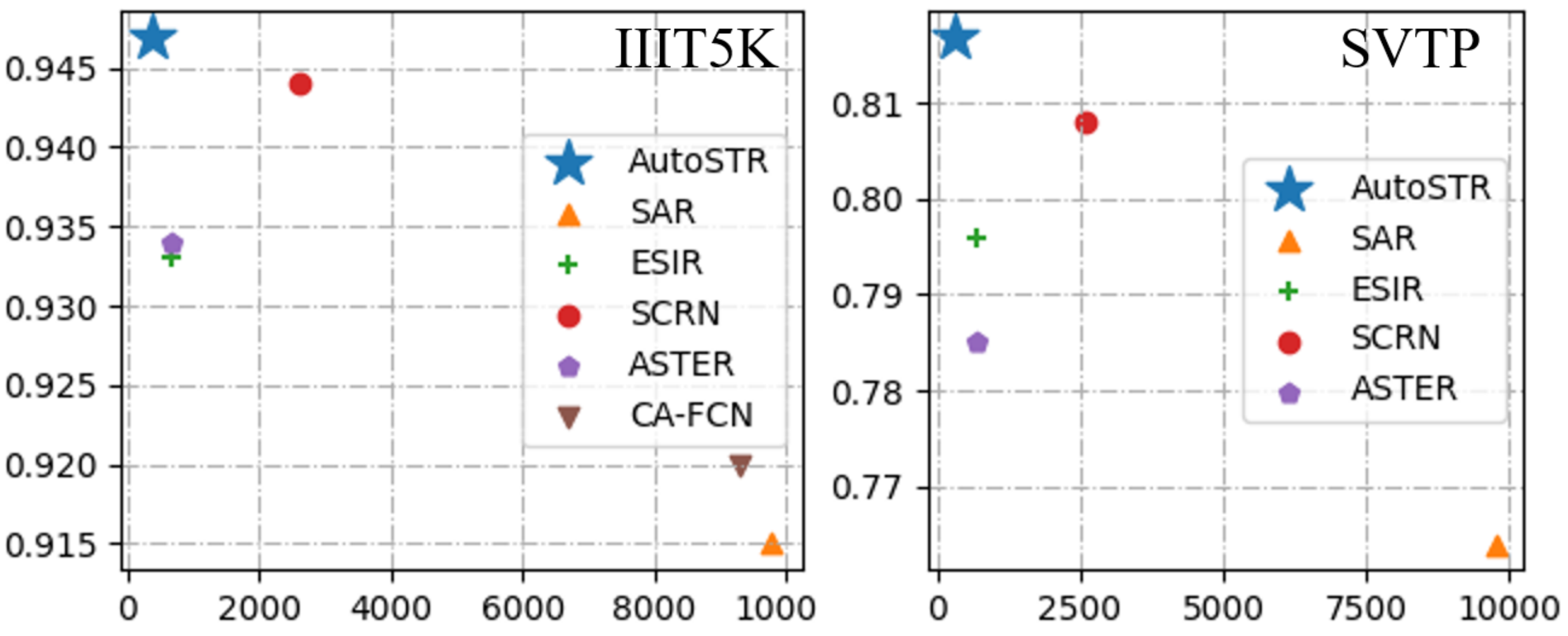}
}
\subfigure[Number of parameters (million).]{\includegraphics[width = 0.45\textwidth]{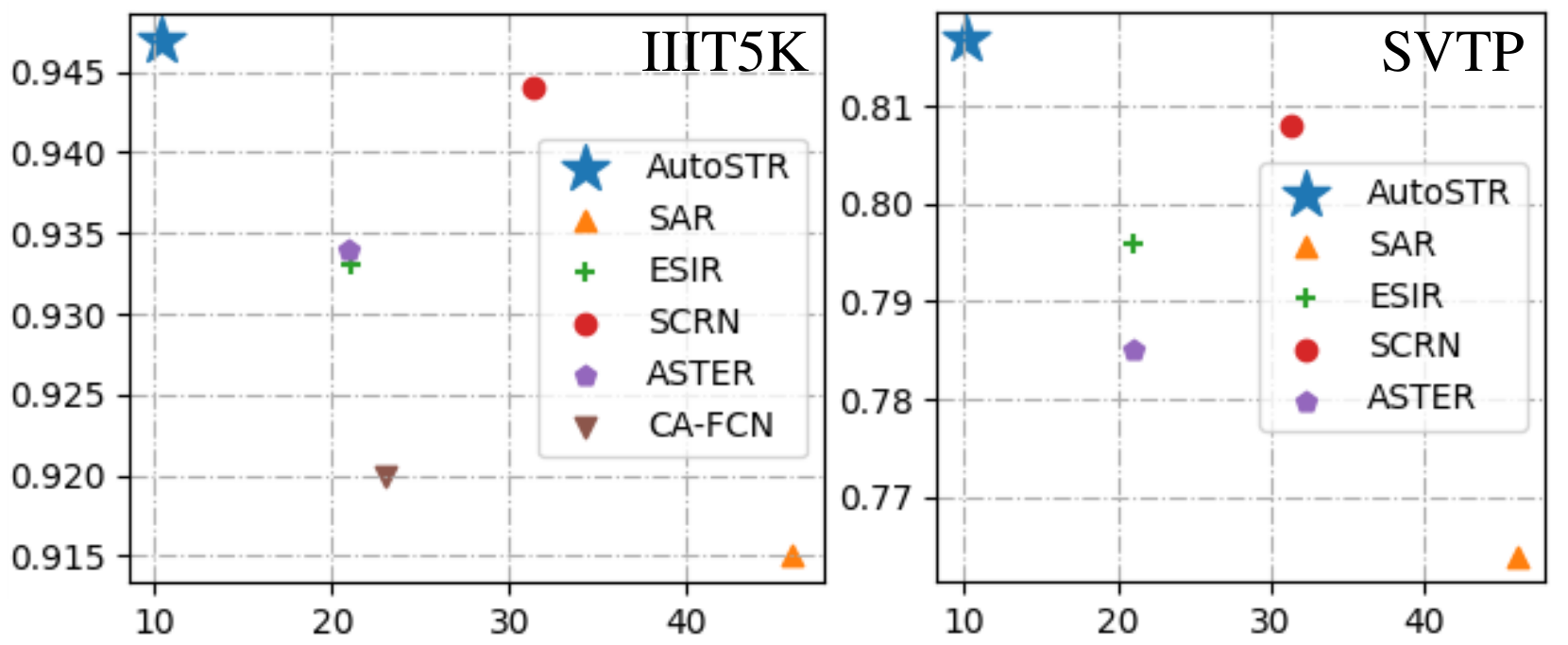}
}
\caption{Accuracy versus computational complexity and memory on IIIT5K and SVTP. 
	Points closer to the top-left are better. 
	Only methods with good performance,
	i.e., with accuracy greater than 90\% on IIIT5K, 
	are plotted.}
\label{fig:overall}
\end{figure}

The comparison on FLOPS and memory size are in 
% \footnote{+qm+ as with the rebuttal,
% 	adding results on one more dataset.
% 	Q2 for Reviewer\#2.}
Fig.\ref{fig:overall}. 
We can see that,
compared with the state-of-the-art methods, 
like SAR~\cite{li2019show}, CA-FCN~\cite{liao2019scene}, ESIR~\cite{zhan2019esir}, SCRN~\cite{yang2019symmetry}, ASTER~\cite{shi2018aster},
the searched architecture cost much less in FLOPS and memory size. 
Thus, 
AutoSTR is much more effective in mobile setting, 
where FLOPS and model size is limited.
%\footnote{+qm+ re-draw the image.}

\subsection{Case Study: Searched Backbones and Discussion}

\subsubsection{Dataset-dependency.}
\begin{figure}[ht]
\centering
\includegraphics[width = 0.84\textwidth]{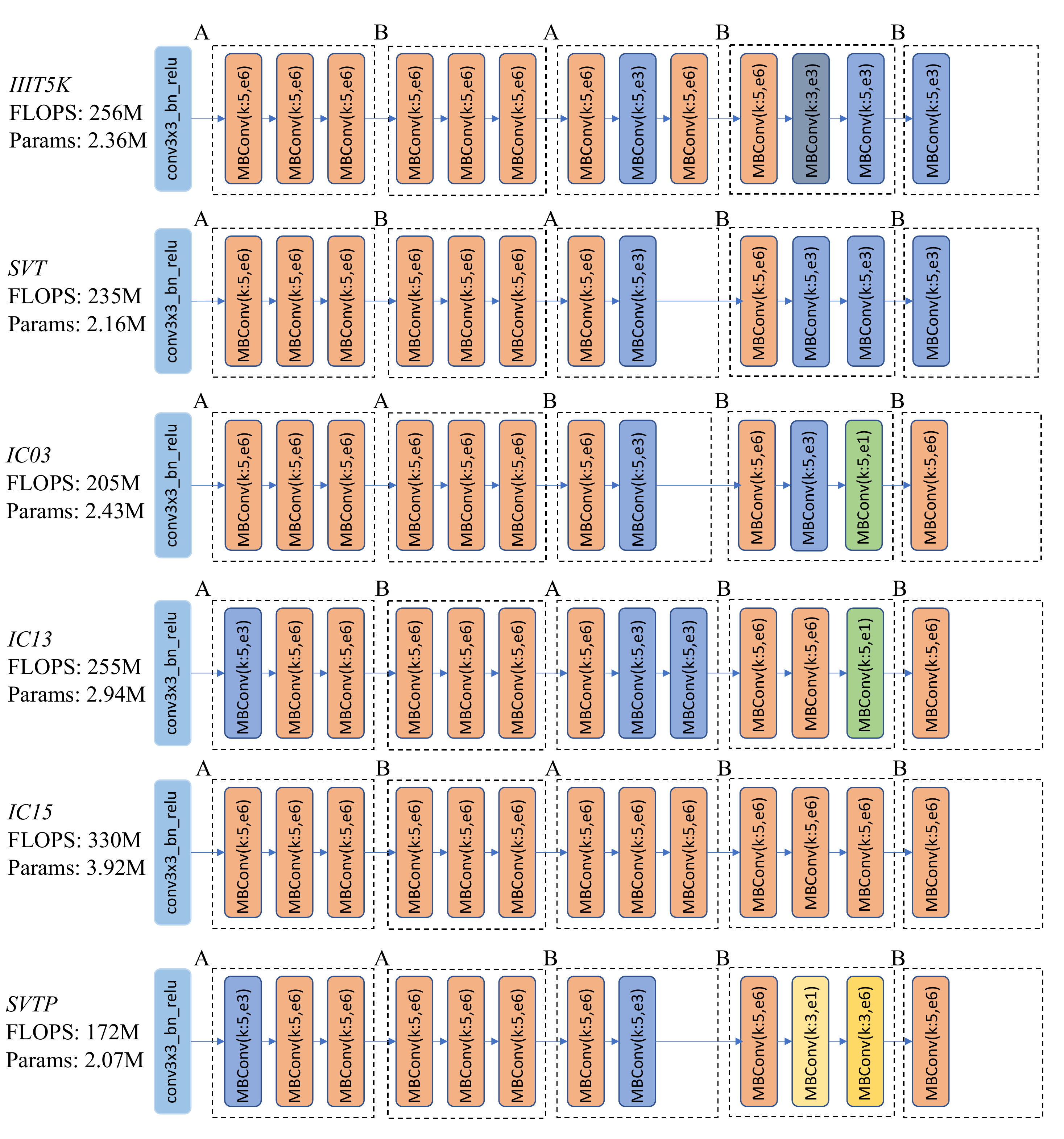}
\includegraphics[width = 0.15\textwidth]{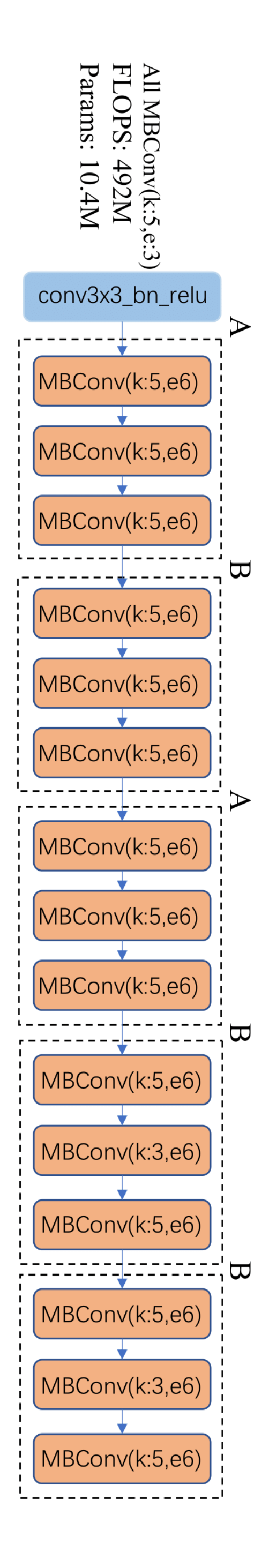}

% \vspace{-12px}
\caption{Left: the searched architectures for AutoSTR in Tab.\ref{tab:main_results}.
	Right: all MBConv(k:5,e:6) baseline architecture.}
\label{fig:searched_archs}
%\label{fig:mb_k5e6_baseline}
\end{figure}

In Fig.\ref{fig:searched_archs}, 
we illustrate architectures of searched feature extractors on each test dataset to give some insights about the design of backbones. 
We observe some interesting patterns. 
The 
shallower convolutional stages (e.g., 1, 2) of the network, prefer larger MBConv operations (e.g., MBConv(k:5,e:6)) 
and do not have skip-connect layer.
But in the deeper convolutional stages (e.g., 3, 4, 5), smaller MBConvs are employed and skip connections are learned to reduce the number of convolutional layers. Especially in the last convolutional stage, only one convolutional layer exists. 
The observed phenomenon is consistent with some manually designed network architecture, such as 
SCRN~\cite{yang2019symmetry}. Specifically, in its first two stages, ResNet50 is used to extract features, and in the later stages, only a few convolutional layers are attached to quickly downsample feature map to generate feature sequences in the horizontal direction.
This phenomenon may inspire us to design better text image feature extractor in the future. 

\subsubsection{Compactness.}
We compare our searched architectures with All MBConv(k:5,e:6) baseline model which choices blocks with the maximum number of parameters in each layer and uses \textsf{ABABB} downsampling strategy as shown in
the right of Fig.\ref{fig:searched_archs}. 
Comparing architectures in Fig.\ref{fig:searched_archs}, we can see that our searched structure has less FLOPS and parameters, while maintain better accuracy, as shown in Tab.\ref{tab:compare_k5e6}.
% \footnote{+++ there is no para and FLOPS in this table} 
Our searched architectures use less FLOPS and parameters, but exceeds the accuracy of the baseline model, 
which explains that the maximum number of parameters model (All MBConv(k:5,e:6) baseline) have lots of redundancy parameters, 
AutoSTR can remove some redundant layers and optimize the network structure.

%\begin{figure}[ht]
%	\centering
%	
%
%	\caption{}
%	
%\end{figure}

\begin{table}[H]
\caption{Accuracies compared with the baseline of All MBConv(k:5,e:6).}
\centering
\setlength\tabcolsep{4pt}
\begin{tabular}{c|c|c|c|c}
\hline
	\textbf{Methods}    & \textbf{IIIT5K} & \textbf{SVT} & \textbf{IC13} & \textbf{IC15}  \\ \hline
	All MBConv(k:5,e:6) & 94.5            & 90.4         & 92.3          & 81.1           \\
	AutoSTR              & 94.7            & 90.9         & 94.2          & 81.8           \\ \hline
\end{tabular}
\label{tab:compare_k5e6}
% \vspace{-10px}
\end{table}

\subsection{Comparison with Other NAS Approaches}

\subsubsection{Search algorithm comparison.}
\begin{figure}[ht]
	\centering
	\includegraphics[width = 0.50\textwidth]{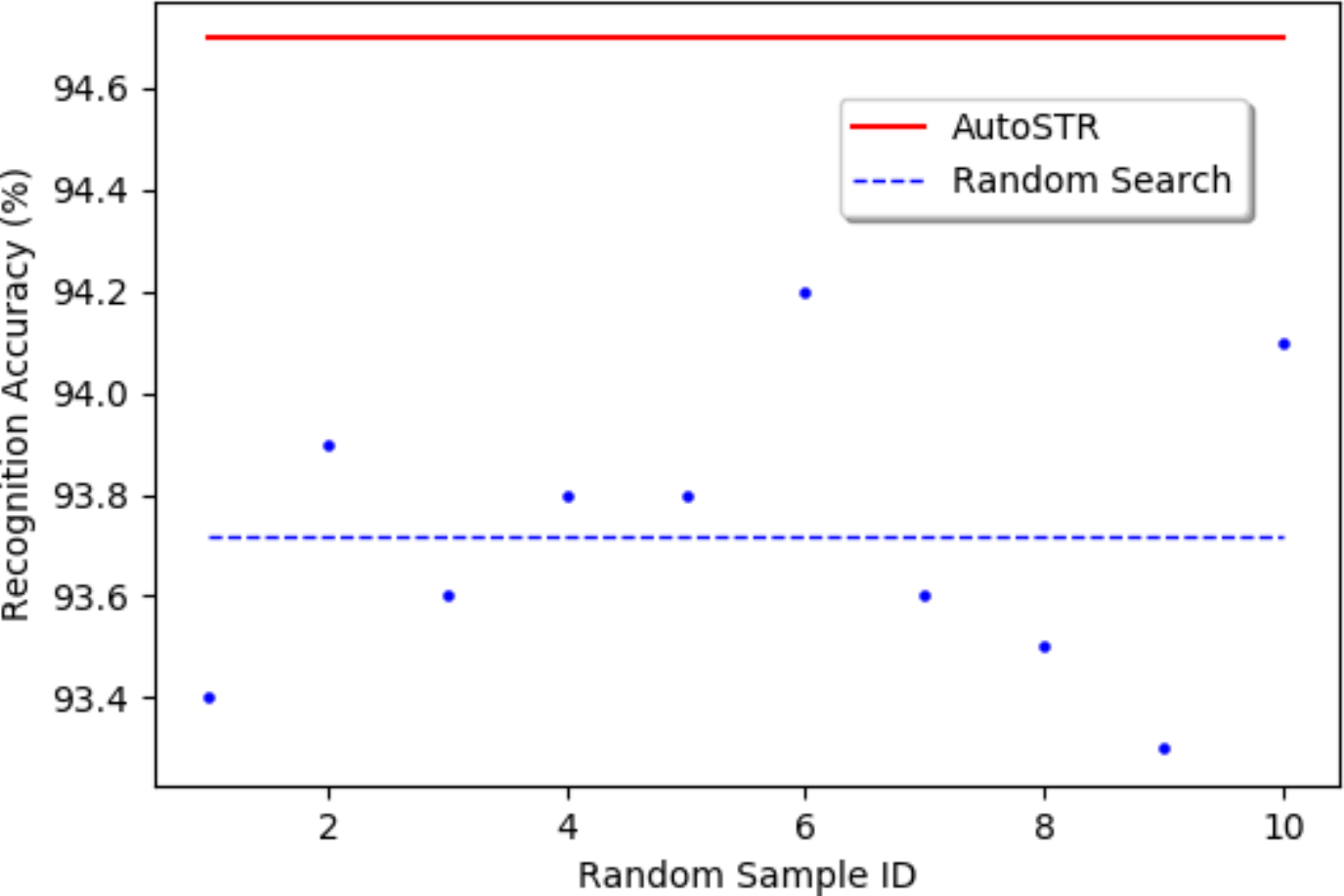}
	\caption{Comparison to random search on IIIT5K dataset.}
	\label{fig:random_search}
\end{figure}

% \footnote{+qm+ statement needs to be changed here,
% 	which should include ``Please see recent benchmarks/surveys [36,23]. Random search is a VERY STRONG baseline''.}
% \footnote{+qm+ what other kinds of algorithms can be compared?}
% As in~\cite{Cai19Proxylessnas,chen2019detnas,sciuto2019evaluating}, 
% random search algorithm is a competitive choice for architecture search. 
% Note that none of other NAS algorithm
% can be used due to the extra search constraint on downsampling path (see Sec~\ref{sec:search_alg}).
From recent benchmarks and surveys \cite{li2019random,chen2019detnas,sciuto2019evaluating}, the random search algorithm is a very strong baseline, we perform a random architecture search from the proposed search space. We choice 10 random architectures, 
train them from scratch, then test these random architectures on IIIT5K dataset. 
Random search takes about $15\times4$GPU days,
while AutoSTR only costs $1.7\times4$GPU days in downsampling-path search step and $0.5\times4$GPU days in operation search step. 
The discovered architecture outperforms random architectures in IIIT5K dataset by 0.5\%-1.4\% as in Fig.\ref{fig:random_search},
which demonstrates AutoSTR is more effectiveness and efficiency.

\subsubsection{Reusing other searched architectures.}
NAS has been extensive researched on image classification task~\cite{DARTS,yao2019differentiable,Cai19Proxylessnas} and segmentation task~\cite{liu2019auto}, et al. 
We study whether those searched architectures are applicable here or not.
Since the constraint in \eqref{eq:opt} cannot be directly satisfied,
we manually tune the searched cell (from DARTS~\cite{DARTS} and AutoDeepLab~\cite{liu2019auto}) in ASTER.
As can be seen from Tab.\ref{tab:com_darts_autodeeplab},
the performance of the backbone from DARTS and AutoDeepLab is much worse.
This further demonstrates direct reusing architecture searched from other tasks is not good.

\begin{table}[ht]
\centering
\setlength\tabcolsep{4pt}
\caption{Comparison with DARTS and AutoDeepLab.}
\begin{tabular}{c|c|c|c|c|c|c}
	\hline
	      \textbf{backbones}       & \textbf{IIIT5K} & \textbf{SVT}  & \textbf{SVT}  & \textbf{IC13} & \textbf{SVTP} & \textbf{IC15} \\ \hline
	  ASTER~\cite{shi2018aster}    &      93.3       &     89.0      &     92.4      &     91.5      &     79.7      &     78.5      \\ \hline
	      DARTS~\cite{DARTS}       &      90.6       &     83.9      &     91.3      &     88.3      &     76.1      &     73.5      \\ \hline
	AutoDeepLab~\cite{liu2019auto} &      93.0       &     87.2      &     91.8      &     91.0      &     77.5      &     76.6      \\ \hline
	           AutoSTR             &  \textbf{94.7}  & \textbf{90.9} & \textbf{93.3} & \textbf{94.2} & \textbf{81.7} & \textbf{81.8} \\ \hline
\end{tabular}
\label{tab:com_darts_autodeeplab}
\end{table}

\subsection{Ablation Study}
\label{sec:abl}

\subsubsection{Downsampling path.}
In our proposed method, we decouple the searching problem in \eqref{eq:opt} into a two-step optimization problem as \eqref{eq:stage1} and \eqref{eq:stage2}. 
This is based on an empirical assumption that a better feature downsampling path can provide a better startup for the operation searching problem, thus can get better architectures easier. 
We use two 
typical strategies in our downsampling path search space, i.e., 
\textsf{AABBB} and 
\textsf{ABABB} to search operations on IIIT5K datasets. 
As shown in Tab.\ref{tab:decouple}, the optimal downsampling path will not be affected by the default operation (i.e. 3$\times$3 residual convolution, MBConv(k:3,e:1)). 
Besides, 
a better downsampling strategy (i.e. \textsf{ABABB}) helps AutoSTR to find a better architecture in the operation search step, which confirms our assumption.

\begin{table}[H]
\centering
\caption{Comparison of different downsampling path on IIIT5K dataset.}
\setlength\tabcolsep{4pt}
\begin{tabular}{c|c|c|c}
\hline
\textbf{Downsample Path}        & \textbf{Default Conv}    & \textbf{Search Step 1} & \textbf{Search Step 2}        \\ \hline
\multirow{2}{*}{\textsf{AABBB}} & 3x3 residual conv        & 92.5           & \multirow{2}{*}{93.9} \\ \cline{2-3}
						& MBConv(k:3,e:1) & 91.3           &                       \\ \hline
\multirow{2}{*}{\textsf{ABABB}} & 3x3 residual onv        & \textbf{93.1}           & \multirow{2}{*}{\textbf{94.7}} \\ \cline{2-3}
						& MBConv(k:3,e:1) & \textbf{92.0}           &                       \\ \hline
\end{tabular}
\label{tab:decouple}
\end{table}

\subsubsection{Impact of the regularizer.} 
In \eqref{eq:reg}, 
we introduce FLOPS into objective function as a regularization term. By adjusting $\beta$, we can achieve the trade off between the calculation complexity and accuracy, 
as shown in Tab.\ref{tab:regularizer}. 

\begin{table}[ht]
\caption{Impact of the regularization on IIIT5K dataset.}
\centering
\setlength\tabcolsep{6pt}
\begin{tabular}{c|c|c|c|c}
	\hline
	$\beta$         & 0.0  & 0.3  & 0.6  & 0.9  \\ \hline
	\textbf{Accuracy (\%)} & 94.6 & 94.5 & 94.7 & 93.5 \\ \hline
	\textbf{FLOPS (M)}   & 319  & 298  & 256  & 149  \\ \hline
	\textbf{Params (M)}   & 3.82 & 3.40 & 2.36 & 1.32 \\ \hline
\end{tabular}
\label{tab:regularizer}
\end{table}

\section{Conclusion}
In this paper, 
we propose to use neural architecture search technology
finding data-dependent sequence feature extraction,
i.e., the backbone,
for the scene text recognition (STR) task.
We first design a novel search space for the STR problem,
which fully explore the prior from such a domain.
Then,
we propose a new two-step algorithm,
which can efficiently search the feature downsampling path and operations separately. 
Experiments demonstrate that our searched backbone can greatly improve the capability of the text recognition pipeline 
and achieve the state-of-the-art results on STR benchmarks. 
As for the future work, we would like to extend the search algorithm to the feature translator.

\section*{Acknowledgments}

The work is performed when H. Zhang was an intern in 4Paradigm Inc. mentored by Dr. Q. Yao.
This work was partially supported by National Key R\&D Program of China (No. 2018YFB1004600), 
to Dr. Xiang Bai by the National Program for Support of Top-notch Young Professionals and the Program for HUST Academic Frontier Youth Team 2017QYTD08.

\clearpage

\bibliographystyle{splncs04}
\bibliography{egbib}

\end{document}